\documentclass[letterpaper, 10 pt, conference]{ieeeconf}  % Comment this line out if you need a4paper

\IEEEoverridecommandlockouts                              % This command is only needed if 
\usepackage{graphics} % for pdf, bitmapped graphics files
\usepackage{graphicx}
\usepackage{amsmath} % assumes amsmath package installed
\usepackage{amssymb}  % assumes amsmath package installed

\usepackage{url}

\let\labelindent\relax
\usepackage{enumitem}

\usepackage{booktabs}
\usepackage{tabularx}
\usepackage{multirow}
\usepackage{multicol}
\usepackage{array}
\usepackage{subcaption}

\newcommand{\se}{\mathfrak{se}}
\newcommand{\SO}{\text{SO}}
\newcommand{\SE}{\text{SE}}

\newcommand{\Exp}{\text{Exp}}
\newcommand{\Log}{\text{Log}}

\usepackage{xcolor}

\DeclareMathOperator*{\argmin}{\arg\!\min}

\newif\ifanonymous
\title{\LARGE \bf
A Comparative Study of Floating-Base Space Parameterizations\\for Agile Whole-Body Motion Planning
}

\ifanonymous
\author{Anonymous Authors$^1$%
\thanks{$^{1}$ Anonymous Affiliation}}
\else
\author{Evangelos Tsiatsianas$^{1,2}$, Chairi Kiourt$^{2,3}$, and Konstantinos Chatzilygeroudis$^{1,2}$% <-this % stops a space
\thanks{*This work has been partially supported by project MIS 5154714 of the National Recovery and Resilience Plan Greece 2.0 funded by the European Union under the NextGenerationEU Program.}% <-this % stops a space
\thanks{$^{1}$Laboratory of Automation and Robotics (LAR) in the Department of Electrical \& Computer Engineering,
        University of Patras, GR-26504 Patras, Greece,
        {\tt\small etsiatsianas@ac.upatras.gr, costashatz@upatras.gr}}%
\thanks{$^{2}$Archimedes/Athena RC, Greece}%
\thanks{$^{3}$Athena - Research and Innovation Center in Information, Communication and Knowledge Technologies, Xanthi, Greece,
        {\tt\small chairiq@athenarc.gr}}%
}
\fi

\begin{document}
\maketitle
\thispagestyle{empty}
\pagestyle{empty}

%%%%%%%%%%%%%%%%%%%%%%%%%%%%%%%%%%%%%%%%%%%%%%%%%%%%%%%%%%%%%%%%%%%%%%%%%%%%%%%%
\begin{abstract}
Automatically generating agile whole-body motions for legged and humanoid robots remains a fundamental challenge in robotics. While numerous trajectory optimization approaches have been proposed, there is no clear guideline on how the choice of floating-base space parameterization affects performance, especially for agile behaviors involving complex contact dynamics. In this paper, we present a comparative study of different parameterizations for direct transcription-based trajectory optimization of agile motions in legged systems. We systematically evaluate several common choices under identical optimization settings to ensure a fair comparison. Furthermore, we introduce a novel formulation based on the tangent space of SE(3) for representing the robot's floating-base pose, which, to our knowledge, has not received attention from the literature. This approach enables the use of mature off-the-shelf numerical solvers without requiring specialized manifold optimization techniques. We hope that our experiments and analysis will provide meaningful insights for selecting the appropriate floating-based representation for agile whole-body motion generation. %Our results show that tangent-space parameterization significantly improve the efficiency and success of complex agile whole-body motion generation for humanoid robots.
% Automatically generating agile whole-body motions remains a fundamental challenge in robotics. While numerous trajectory optimization approaches have been proposed, there is currently no clear guideline on which parameterization to select for different tasks or settings. In this work, we conduct a comprehensive evaluation of several common choices for direct transcription-based trajectory optimization of agile whole-body motions. To ensure a fair comparison, all methods are evaluated using the same optimizer and numerical settings. Additionally, we introduce a transcription strategy based on the tangent space of SE(3) for representing the robot’s base pose — a parameterization that, to our knowledge, has not been widely adopted in previous work. This formulation enables the use of mature off-the-shelf numerical solvers without requiring custom manifold optimization techniques. Our results show that working in the tangent space of SE(3) leads to more efficient and effective generation of agile whole-body motions compared to other commonly used parameterizations.
\end{abstract}
%%%%%%%%%%%%%%%%%%%%%%%%%%%%%%%%%%%%%%%%%%%%%%%%%%%%%%%%%%%%%%%%%%%%%%%%%%%%%%%%
\section{Introduction}\label{sec:intro}
Generating agile whole-body motions for legged and humanoid robots is a fundamental capability for enabling dynamic and versatile behaviors such as jumping, running, or fast reactivity to external disturbances~\cite{koolen2016design,winkler2018gait,tsikelis2024gait}. Trajectory optimization has emerged as a powerful framework for planning such motions by directly computing physically consistent trajectories that respect the robot's dynamics and contact constraints~\cite{posa2014direct,kuindersma2016optimization,wensing2023optimization}. However, the choice of floating-base space parameterization --- that is, how robot floating-base poses are represented during optimization --- plays a critical yet often overlooked role in determining the efficiency, robustness, and success of the resulting trajectories. Despite the variety of parameterizations proposed in the literature, there is currently no systematic evaluation guiding the selection of an appropriate representation for agile whole-body motion planning. In this work, we aim to bridge this gap by conducting a comparative study of different floating-base space parameterizations for direct transcription-based trajectory optimization, highlighting their impact on the generation of agile motions in legged systems.

Prior work has explored various representations such as Euler angles~\cite{winkler2018gait}, quaternions~\cite{jackson2021planning}, or full SE(3) transformations~\cite{jallet2025proxddp}, but typically focuses on a single formulation without direct comparison to alternatives. It is clear from the literature that operating in the $\SE(3)$ manifold directly should yield better solutions and more robust procedures~\cite{barfoot2024state,sola2018micro,jallet2025proxddp,tsikelis2025multi}. Nevertheless, approaches that rely on manifold-based formulations often require specialized solvers or custom handling of constraints, which can limit practical deployment~\cite{jallet2025proxddp,teng2025riemannian}.
% In contrast, parameterizations that operate in locally linear spaces, such as tangent spaces, offer the potential for leveraging mature off-the-shelf numerical optimization libraries~\cite{teng2025riemannian}.
In this manuscript, we make the case that operating in the tangent space of $\SE(3)$ allows for the usage of mature and robust numerical solvers (e.g., Ipopt~\cite{wachter2006_IPOPT}) while maintaining sufficient expressiveness for agile behaviors. This observation motivates our comprehensive evaluation of different floating-base space choices for agile whole-body motion planning in legged systems.

\textbf{The main contributions of this work are:}
\begin{itemize}
    \item We present a comparative study of different floating-base space parameterizations for direct transcription-based trajectory optimization of agile whole-body motions in legged and humanoid robots.
    \item We systematically evaluate commonly used representations under identical optimization and numerical settings to ensure fairness.
    \item We introduce a novel trajectory transcription based on the tangent space of SE(3) for representing the robot's floating-base pose, enabling the use of mature numerical solvers without specialized manifold optimization.
    % \item We provide extensive experimental validation on a set of agile motion generation tasks, showing that parameterizations based on the SE(3) manifold significantly improve optimization efficiency and motion quality compared to conventional alternatives.
\end{itemize}

\begin{figure*}
    \centering
    \includegraphics[width=\linewidth]{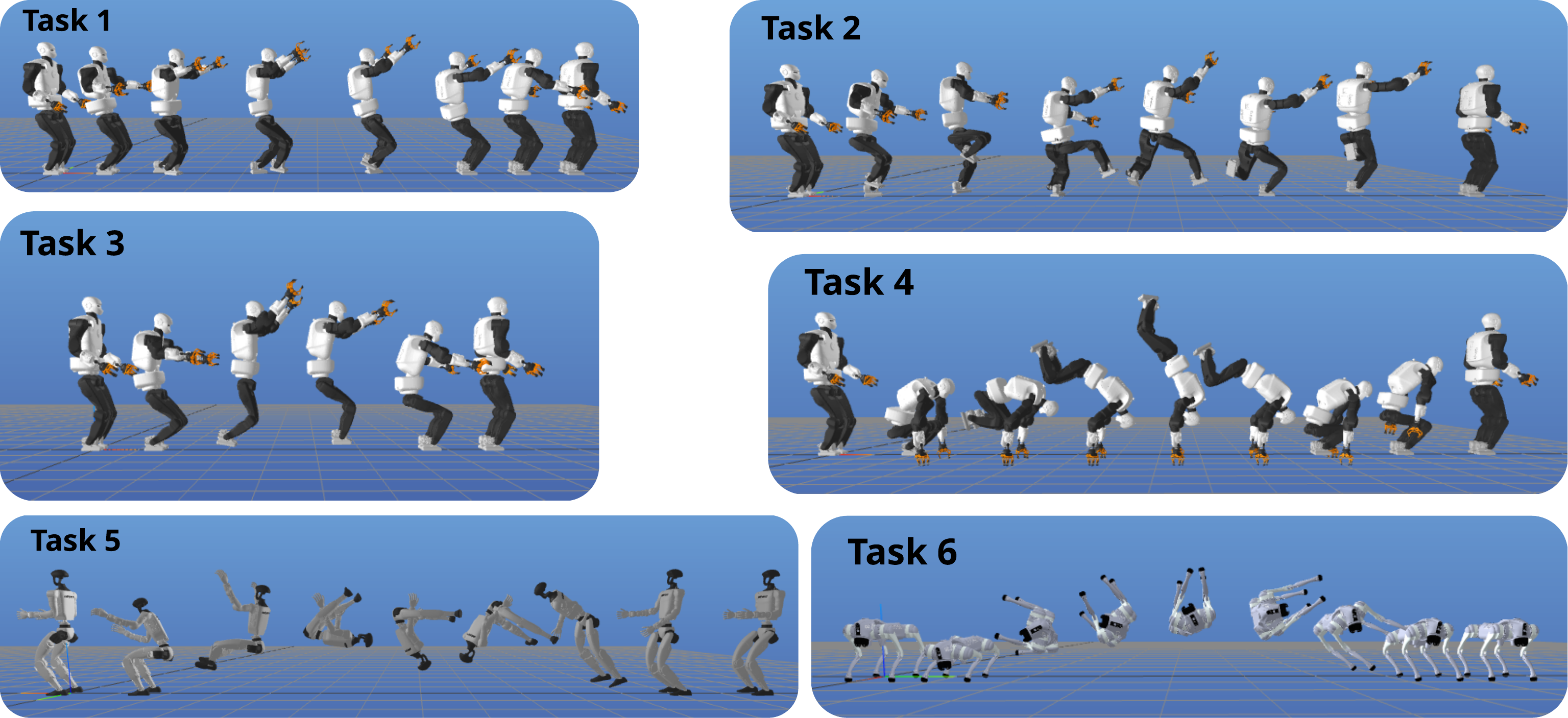}
    \caption{Overview of the different tasks used for comparisons and indicative solutions. \textbf{Task 1 (Walk):} a Talos humanoid walks $2\,\mathrm{m}$ forward. \textbf{Task 2 (Hopscotch):} a Talos humanoid is hopping first on the left leg and then on the right ($2\,\mathrm{m}$ forward). \textbf{Task 3 (Big jump):} a Talos humanoid is jumping $1\,\mathrm{m}$ forward. \textbf{Task 4 (Handstand):} a Talos humanoid is performing a handstand in place. \textbf{Task 5 (Humanoid back-flip):} a Unitree G1 humanoid is performing a back-flip (it has to land $0.5\,\mathrm{m}$ backward). \textbf{Task 6 (Quadruped side-flip):} a Unitree Go2 quadruped is peforming a side-flip  (it has to land $0.3\,\mathrm{m}$ sideways). \emph{For visualization purposes, we have added space between the timesteps}. Please refer to the supplementary video for the real-time visualizations.}
    \label{fig:overview}
    \vspace{-1em}
\end{figure*}
\section{Related Work}\label{sec:related}
The choice of the robot's floating-base representation directly influences the transcription of the trajectory optimization problem, which in turn can affect both solution quality and convergence speed. Early methods parameterized orientation via Euler angles, which offer intuitive roll–pitch–yaw decomposition but suffer from poor numerical conditioning when approaching the gimbal-lock singularities. Switching to unit quaternions removes singularities and provides smooth interpolation, yet introduces a non-minimal 4D representation with a unit-norm constraint and a double-cover ambiguity that must be carefully enforced throughout optimization. Operating directly on the $\SE(3)$ manifold avoids these issues, but requires specialized manifold-aware solvers to maintain group structure~\cite{barfoot2024state,sola2018micro,jallet2025proxddp}.

Despite these varied proposals, prior work typically evaluates only one representation in isolation --- Euler angles~\cite{winkler2018gait}, quaternions~\cite{jackson2021planning}, or $\SE(3)$~\cite{jallet2025proxddp} --- without a head-to-head comparison under identical settings. Consequently, there is no clear guideline for selecting the most suitable floating-base space for agile behaviors involving fast dynamics and complex contacts.

Meanwhile, the numerical optimization community has developed robust, high-performance Non-Linear Programming (NLP) solvers that operate in Euclidean vector spaces, demonstrating reliable convergence across large-scale diverse problems (including robotic applications and trajectory optimization~\cite{wensing2023optimization}). However, these solvers do not natively support manifold constraints, forcing most implementations to fall back on vectorized parameterizations (e.g. quaternions).

In contrast, manifold optimization techniques for $\SE(3)$ and other Lie groups~\cite{absil2008optimization} have only recently gained traction in robotics and numerical optimal control. While promising algorithms (e.g., ProxDDP~\cite{jallet2025proxddp} or Riemannian direct methods~\cite{teng2025riemannian}) show improved theoretical appeal, they rely on custom implementations, and we are yet to see a robust implementation widely adopted and tested by the community. Overall, NLP solvers that operate in manifolds are still in early stage, and do not provide the robustness and effectiveness of the solvers that operate in vector spaces.

In this work, we bridge these two streams by showing that one can enjoy the expressiveness of $\SE(3)$ while still leveraging off-the-shelf vector-space NLP solvers. By formulating floating-base poses in the tangent space of $\SE(3)$, we maintain manifold consistency without sacrificing solver robustness. We systematically evaluate this approach against Euler Angles and Quaternions for agile whole-body motion planning in legged systems.
\section{Background}\label{sec:bg}
Planning agile whole-body motions for legged robots is a challenging task, and can be tackled in many ways. In particular, trajectory optimization has emerged as a powerful framework for generating dynamically consistent motions~\cite{wensing2023optimization}, relying heavily on accurate models of whole-body dynamics and suitable representations of the robot’s state. This section outlines the mathematical foundations underlying our study.
% beginning with an overview of trajectory optimization as a method for solving continuous-time optimal control problems. We then introduce the equations of motion for floating-base systems, followed by a detailed examination of two widely used representations for rigid-body pose: elements of the Lie group SE(3) and unit quaternions. These formulations form the basis for the floating-base space parameterizations compared in this work.
\subsection{Special Euclidean Group}\label{sec:se3}
The Special Euclidean group $\SE(3)$ describes rigid-body motions, combining translations and rotations, and is endowed with a Lie group structure. A rigid-body pose $\boldsymbol{T}\in\SE(3)$ is defined as:
\begin{align}
    \boldsymbol{T} = \begin{bmatrix}\boldsymbol{R} & \boldsymbol{p} \\ \boldsymbol{0} & 1\end{bmatrix} \in\mathbb{R}^{4\times4},
\end{align}
where $\boldsymbol{R}\in\SO(3)$ is a rotation matrix in the Special Orthogonal group, and $\boldsymbol{p}\in\mathbb{R}^3$ is a translation vector.

The \textbf{tangent space at the identity}, denoted $\se(3)$, is the associated Lie algebra and provides a local linear approximation of the manifold~\cite{sola2018micro,barfoot2024state}. Although elements of $\se(3)$ have a non-trivial matrix structure, they can be expressed as linear combinations of basis elements. Hence, they are commonly represented via their vector coordinates, $ \boldsymbol{\xi}\in\mathbb{R}^6$.% The corresponding vector in the tangent space is given by:
% \begin{align*}
%     \boldsymbol{\xi} = \begin{bmatrix}\boldsymbol{\rho} \\ \boldsymbol{\theta}\end{bmatrix} \in \mathbb{R}^6,
% \end{align*}
% where $\boldsymbol{\rho}$ and $\boldsymbol{\theta}$ represent translational and rotational components, respectively.

The exponential and logarithmic maps provide a means to move between the manifold and its tangent space (we use the notation from~\cite{sola2018micro}):
\begin{align}
    \Exp(\boldsymbol{\xi}) &= \boldsymbol{T},\nonumber\\
    \Log(\boldsymbol{T}) &= \boldsymbol{\xi}.
\end{align}
Given a small increment $\Delta\boldsymbol{\xi} = \mathcal{V}_bh\in\mathbb{R}^6$ (where $\mathcal{V}_b$ is the body twist and $h$ is the timestep), the retraction operation in $\SE(3)$ is defined as\footnote{We use $k$ as the time (or knot point) index.}:
\begin{align}
     \boldsymbol{T}_{k+1} =  \boldsymbol{T}_k \oplus\Delta\boldsymbol{\xi} =  \boldsymbol{T}_k\Exp(\Delta\boldsymbol{\xi}),
\end{align}
which ensures that $ \boldsymbol{T}_{k+1}$ remains on the manifold. Similarly, we get:
\begin{align}
     \Delta\boldsymbol{\xi} = \boldsymbol{T}_{k+1}\ominus\boldsymbol{T}_k  = \Log(\boldsymbol{T}_k^{-1}\boldsymbol{T}_{k+1}).
\end{align}
Using the tangent space of $\SE(3)$ enables efficient differentiation~\cite{sola2018micro,barfoot2024state}. In particular, the left and right Jacobians linearly map local perturbations (e.g., body twists) to variations on the manifold. These Jacobians obey the chain rule for Lie groups~\cite{sola2018micro} and are crucial for computing gradients during trajectory optimization.
Ultimately, the tangent space representation of $\SE(3)$ provides a geometrically consistent and computationally efficient framework for the floating-based representation.% It naturally enforces rigid-body constraints, avoids singularities, and facilitates the seamless integration of pose optimization within whole-body dynamic models.
\subsection{Euler Angles}\label{sec:euler_angles}
An alternative to $\SE(3)$ is the use of Euler Angles for encoding orientation, and a vector $\boldsymbol{p}\in\mathbb{R}^3$ for the translation. We define the \emph{roll} \(\phi\), \emph{pitch} \(\theta\), and \emph{yaw} \(\psi\) as successive rotations about the body‐fixed axes \(x\), \(y\), and \(z\), respectively. We denote this as $\boldsymbol{\theta}=\begin{bmatrix}\phi & \theta & \psi\end{bmatrix}^\top\in\mathbb{R}^3$. The complete rotation from the inertial/world frame to the body frame is:
\begin{align}
    \boldsymbol{R}(\phi,\theta,\psi) \;=\; \boldsymbol{R}_z(\psi)\,\boldsymbol{R}_y(\theta)\,\boldsymbol{R}_x(\phi),
\end{align}

where $\boldsymbol{R}_z,\boldsymbol{R}_y,\boldsymbol{R}_x$ define the rotation matrices around the principle axes. We can relate an angular velocity $\boldsymbol{\omega}_b \in \mathbb{R}^3$ expressed in the body frame, with the roll-pitch-yaw parameters time derivatives as follows:
\begin{align}\label{eq:rpy_integration}
    \dot{\boldsymbol{\theta}}
    = \underbrace{\begin{bmatrix}
    1 & \sin\phi\tan\theta    & \cos\phi\tan\theta \\
    0 & \cos\phi               & -\sin\phi          \\
    0 & \sin\phi/\cos\theta    & \cos\phi/\cos\theta
    \end{bmatrix}}_{\boldsymbol{W}(\boldsymbol{\theta})}
    \boldsymbol{\omega}_b.
\end{align}
This formulation enables integration using standard numerical schemes (e.g., Euler or Runge-Kutta). However, using this representation, singularities occur when \(\theta=\pm 90^\circ\), known as \emph{gimbal lock}. Overall, this representation offers an intuitive, human-readable description of orientation by directly relating rotations to familiar notions of roll, pitch, and yaw, but singularities can be difficult to handle generically.
\subsection{Quaternions}\label{sec:quats}
Another alternative representation is the use of unit quaternions for encoding orientation, and a vector for translation. Unit quaternions $\boldsymbol{\rho}\in \mathbb{H}$ form a globally non-singular, smooth representation of 3D rotations. We denote a quaternion as:
\begin{align}
    \label{eq:quat}
    \boldsymbol{\rho} = \begin{bmatrix} s\\\boldsymbol{\nu} \end{bmatrix} = 
    \begin{bmatrix}
        \cos\frac{\theta}{2}\\
        \boldsymbol{r} \sin\frac{\theta}{2}
    \end{bmatrix} \in \mathbb{R}^4,
\end{align}
where $\boldsymbol{\nu} \in \mathbb{R}^3$ is the vector part, $s \in \mathbb{R}$ is the scalar part, and $\boldsymbol{r}$ and $\theta$ define the axis and angle of rotation, respectively. From Eq.~\eqref{eq:quat} it follows that a quaternion that describes a rotation needs to be of unit norm, $\|\boldsymbol{\rho}\| = 1$.

We can relate an angular velocity $\boldsymbol{\omega}_b \in \mathbb{R}^3$ expressed in the body frame, with the quaternion parameters time derivatives as follows~\cite{jackson2021planning}:
\begin{align}
    \label{eq:quat_integration}
    \dot{\boldsymbol{\rho}} = \frac{1}{2}\boldsymbol{L}(\boldsymbol{\rho}) \boldsymbol{H}\boldsymbol{\omega}_b,
\end{align}
where
\begin{align*}
    \boldsymbol{L}(\boldsymbol{\rho}) &= \begin{bmatrix}
        s & -\boldsymbol{\nu}^\top \\
        \boldsymbol{\nu} & s\boldsymbol{I}_3 + \hat{\boldsymbol{\nu}}
    \end{bmatrix}, \qquad
    \boldsymbol{H} = \begin{bmatrix}
        \boldsymbol{0}_{1\times3}\\\boldsymbol{I}_3
    \end{bmatrix}.
\end{align*}
This formulation enables quaternion integration using standard numerical schemes.
% However, such integration may violate the unit-norm constraint, necessitating explicit normalization during trajectory optimization.
%
Quaternions eliminate singularities and enable smooth interpolation, avoiding issues such as gimbal lock. Nonetheless, their non-minimal 4D representation introduces a unit-norm constraint (or normalization\footnote{In our implementation, we always normalize the quaternions and take the appropriate derivatives.}) that must be enforced throughout optimization. Additionally, the double-cover property of quaternions (i.e., \(\boldsymbol{\rho}\equiv-\boldsymbol{\rho}\)) requires careful handling to maintain derivative consistency in the objective and constraint functions.
\subsection{Optimal Control Problem \& Trajectory optimization}\label{sec:trajopt}
The generic continuous-time optimal control problem is given by:
\begin{align*}
    \min_{\boldsymbol{x}(t), \boldsymbol{u}(t)} \quad & \mathcal{J}(x(t), u(t)) = \int_{t_0}^{t_f}  \ell(\boldsymbol{x}(t), \boldsymbol{u}(t)) \, dt + \ell_f(\boldsymbol{x}(t_f)) \\
    \text{s.t.} \quad & \dot{\boldsymbol{x}}(t) = f(\boldsymbol{x}(t), \boldsymbol{u}(t))
\end{align*}
where: 
\begin{itemize}
    \item \(\boldsymbol{x}(t), \boldsymbol{u}(t) \) are the state and control trajectories,
    \item \(\dot{\boldsymbol{x}}(t)\) the state's time derivative
    \item $\ell$ is the "stage cost" with $\ell_f$ defining the "terminal cost".
\end{itemize}

The above definition leads to an infinite dimension problem. There are several methods for transcribing it into a nonlinear program of finite dimension suitable for numerical optimization. In this work, we employ \emph{inverse dynamics direct transcription}~\cite{ferrolho2021inverse}, where the continuous-time problem is discretized as:
\begin{align}
    \min_{\boldsymbol{x}_{0:N}, \dot{\boldsymbol{x}}_{0:N-1}, \boldsymbol{u}_{0:N-1}} \quad & J = \sum_{k=0}^{N-1} \boldsymbol{\ell}(\boldsymbol{x}_k, \boldsymbol{u}_k) + \boldsymbol{\ell}_f(\boldsymbol{x}_N) \\
    \text{s.t.} \quad & \dot{\boldsymbol{x}}_k = f(\boldsymbol{x}_k, \boldsymbol{u}_k)\nonumber\\
    & \boldsymbol{x}_{k+1} = \text{integrate}(\boldsymbol{x}_k, \dot{\boldsymbol{x}}_k)\nonumber
\end{align}
\subsection{Whole-Body Dynamics Formulation}\label{sec:dynamics}

We model robots as free-floating multibody systems, composed of a rigid base and $n$ actuated joints\footnote{For simplicity, we assume that the joints operate in a vector space, although this is not necessarily the case (e.g. a spherical joint). We can adapt the formulations using composite manifolds~\cite{sola2018micro}.}. The generalized configuration of the robot is denoted by $\boldsymbol{q} \in \mathcal{Q} = \mathcal{G} \times \mathbb{R}^n$, where $\mathcal{G}$ encodes the pose of the floating base, and $\mathbb{R}^n$ corresponds to the internal joint configuration. The generalized velocities of the robot are denoted by $\boldsymbol{v}\in\mathcal{H} = \mathcal{V}\times\mathbb{R}^n$, where $\mathcal{V}$ encodes the spatial twist of the floating-base~\cite{featherstone2014,lynch2017modern}, and $\mathbb{R}^n$ corresponds to the velocities of the joints. The system dynamics are described by the manipulator equation~\cite{featherstone2014}:

\begin{equation}\label{eq:eom}
    \boldsymbol{M}(\boldsymbol{q})\dot{\boldsymbol{v}} + \boldsymbol{C}(\boldsymbol{q}, \boldsymbol{v}) = \boldsymbol{S}\boldsymbol{\tau} + \sum_{i=1}^{k} \boldsymbol{J}_i^\top(\boldsymbol{q})\boldsymbol{\lambda}_i,
\end{equation}
where $\boldsymbol{M}(\boldsymbol{q})$ denotes the symmetric positive-definite generalized inertia matrix. The term $\boldsymbol{C}(\boldsymbol{q}, \boldsymbol{v})$ represents Coriolis, centrifugal effects and gravitational forces. The actuation selection matrix $\boldsymbol{S} = \begin{bmatrix}\boldsymbol{0} & \boldsymbol{I}\end{bmatrix}^\top$ maps joint torques $\boldsymbol{\tau} \in \mathbb{R}^n$ into the full configuration space. Contact interactions are described via the Jacobian matrices $\boldsymbol{J}_i(\boldsymbol{q})$, which relate joint space to spatial velocities at contact points, and the corresponding contact wrenches $\boldsymbol{\lambda}_i \in \mathbb{R}^6$.

The choice of representation for the floating-base configuration directly influences the dimension and structure of the configuration space $\mathcal{G}$, and consequently the generalized coordinate vector $\boldsymbol{q}$. In this manuscript, we consider three alternative representations. The first is the tangent space of the Special Euclidean group at the identity, $\mathcal{G} = \se(3)$, which provides a minimal, locally linearized representation of the base pose. The second uses unit quaternions for orientation, combined with Euclidean coordinates for translation, yielding $\mathcal{G} = \mathbb{H} \times \mathbb{R}^3$. Lastly, we consider using Euler-Angles for (absolute) orientation along with Euclidean coordinates for translation, yielding $\mathcal{G} = \mathbb{R}^6$. Each of these choices impacts both the formulation of the equations of motion and the numerical properties of the trajectory optimization.
%
% \vspace{-1.5em}
\section{Trajectory Optimization Formulations}\label{sec:spaces}
To compute dynamically consistent whole-body motions, we formulate a nonlinear trajectory optimization problem by transcribing the system's differential equations into algebraic constraints using direct collocation. The optimization yields trajectories that respect the robot’s full-body dynamics, contact feasibility, and kinematic constraints.

The decision variables at each discretization node \( k = 0, \dots, N \) include the robot's generalized configuration \( \boldsymbol{q}_k \), generalized velocities \( \boldsymbol{v}_k \), and accelerations \( \dot{\boldsymbol{v}}_k \), as well as the contact forces (and not wrenches\footnote{We use multiple contact points per contact frame when necessary (e.g. we use 4 contact points for rectangular feet).}) \( \boldsymbol{\lambda}_{j,k} \in \mathbb{R}^3 \) and contact positions \( \boldsymbol{c}_{j,k} \in \mathbb{R}^3 \) for each foot \( j \) in contact at time step \( k \).

The contact schedule is predefined and encoded via the index set \( \mathcal{C}_k \subseteq \{1, \dots, n_f\} \), which indicates the frames in contact at each stage. The integration time step \( h \) is kept constant throughout the horizon. Preliminary attempts to include \( h \) as a decision variable did not improve performance or solution quality, and are therefore omitted in the final formulation.

% To compute a feasible motion plan, we transcribe the differential equations to algebraic equations and solve them as presented in section \ref{sec:dynamics}. The optimization problem we formulate contains as decision variables the robot's generalized coordinates $\boldsymbol{q}$, their time derivatives $\dot{\boldsymbol{q}}$, $\ddot{\boldsymbol{q}}$ and contact positions $\boldsymbol{c}_j$ with contact forces $\boldsymbol{\lambda}_j$ for each foot in contact at the determined stage. The integration timestep between nodes is denoted as $\boldsymbol{h}$, although step size was tested as an optimization variable, it didn't yield any adequately results.
% \todo{contact explicit reference, predefined gait}\\
The transcribed numerical optimization is as follows:
\begin{align}
    \argmin_{\substack{
        \boldsymbol{q}_k, \boldsymbol{v}_k, \dot{\boldsymbol{v}}_k \\
        \boldsymbol{\lambda}_{j,k}, \boldsymbol{c}_{j,k}
    }}
    J &= \sum_{k=0}^{N-1} \ell(\boldsymbol{q}_k, \boldsymbol{v}_k, \dot{\boldsymbol{v}}_k, [\boldsymbol{\lambda}]_k) + \ell_f(\boldsymbol{q}_N, \boldsymbol{v}_N)\\
    \text{s.t.} \quad 
    & \begin{bmatrix}\boldsymbol{0}_{6} \\ \boldsymbol{\tau}_{\min}\end{bmatrix}\leq\boldsymbol{\tau}(\boldsymbol{q}_k, \boldsymbol{v}_k, \dot{\boldsymbol{v}}_k,[\boldsymbol{\lambda}]_k)\leq\begin{bmatrix}\boldsymbol{0}_{6} \\ \boldsymbol{\tau}_{\max}\end{bmatrix}\\% && \textit{(Inverse Dynamics)}\\ 
    %add a matrix for integration (abstract of maths)
    & \text{integrate}(\boldsymbol{q}_k, \boldsymbol{v}_k, \dot{\boldsymbol{v}}_k, \boldsymbol{q}_{k+1}, \boldsymbol{v}_{k+1}, h) = \boldsymbol{0}\\% && \textit{(Time Integration)}\\
    % &   \boldsymbol{q}_{k+1} = \boldsymbol{q}_k \oplus \dot{\boldsymbol{q}}_k dt & \textit{(Time Integration)} \\
    % &   \dot{\boldsymbol{q}}_{k+1} = \dot{\boldsymbol{q}}_k + \ddot{\boldsymbol{q}}_k dt \\
    & \text{FK}_j(\boldsymbol{q}_k) = c_{j,k} \quad j\in\mathcal{C}_k\\% && \textit{(Contact Kinematics)} \\
    & \dot{\boldsymbol{c}}_{j,k} = 0\\% && \textit{(Contact Constraints)} \\
    & \boldsymbol{c}_{j,k}\in\text{Contact Region} \\
    & \boldsymbol{q}_k \in \mathcal{Q}, \quad \boldsymbol{v}_k \in \mathcal{H}\\% && \textit{(Kinematic Limits)} \\
    &   \boldsymbol{\lambda}_{j,k}\in\Lambda% && \textit{(Friction Cones)} \\
\end{align}  
where $\boldsymbol{\tau}(\boldsymbol{q}_k, \boldsymbol{v}_k, \dot{\boldsymbol{v}}_k,[\boldsymbol{\lambda}]_k)$ refers to solving Eq.~\eqref{eq:eom} for $\boldsymbol{\tau}$, $\boldsymbol{\tau}_{\text{min}}$ and $\boldsymbol{\tau}_{\text{max}}$ are the joint torque limits, $\text{integrate}(\cdot,\cdot,\cdot,\cdot,\cdot,\cdot)$ defines a numerical integration scheme, $\text{FK}_j(\cdot)$ performs forward kinematics for the $j$-th frame, and $\Lambda$ define the linearized (pyramid) friction cone constraints~\cite{wensing2023optimization,chatzilygeroudis2023evolving}.

\begin{table}[!htbp]
  \centering
  % \small
  \caption{Floating-base Space Representations and Operations}
  \label{tab:kinematic-ops}
  \begin{tabularx}{\linewidth}{@{} %l
                                % >{\centering\arraybackslash}X 
                                >{\centering\arraybackslash}p{0.15\linewidth}
                                >{\centering\arraybackslash}p{0.1\linewidth}
                                % >{\centering\arraybackslash}X 
                                % p{0.15\linewidth}
                                >{\centering\arraybackslash}p{0.39\linewidth}
                                >{\centering\arraybackslash}X @{}}
    \toprule
    Name          & Variables 
                  & Differences 
                  & Elementary Integration \\
    \midrule
    \textbf{SE(3) Tangent} & $\boldsymbol{\xi}_k$     
                  & $\Exp(\boldsymbol{\xi}_2)\!\ominus\!\Exp(\boldsymbol{\xi}_1)$ 
                  & $\Log(\Exp(\boldsymbol{\xi})\oplus\mathcal{V}_bh)$ \\
    \midrule
    \textbf{Quaternions \#1} & $\begin{bmatrix}\boldsymbol{p}_k\\\boldsymbol{\rho}_k\end{bmatrix}$
                  & $\begin{bmatrix}\boldsymbol{p}_2-\boldsymbol{p}_1\\\boldsymbol{\rho}_2-\boldsymbol{\rho}_1\end{bmatrix}$
                  & $\begin{bmatrix}\boldsymbol{p}+\dot{\boldsymbol{p}}h\\\boldsymbol{\rho}+\frac{1}{2}\boldsymbol{L}(\boldsymbol{\rho}) \boldsymbol{H}\boldsymbol{\omega}_bh\end{bmatrix}$ \\
    \midrule
    \textbf{Quaternions \#2} & $\begin{bmatrix}\boldsymbol{p}_k\\\boldsymbol{\rho}_k\end{bmatrix}$
                  & $\begin{bmatrix}\boldsymbol{p}_2-\boldsymbol{p}_1\\\boldsymbol{\rho}_2-\boldsymbol{\rho}_1\end{bmatrix}$
                  & $\Log_{q}(\Exp_{q}(\begin{bmatrix}\boldsymbol{p}\\\boldsymbol{\rho}\end{bmatrix})\oplus\mathcal{V}_bh)$ \\
    \midrule
    \textbf{Quaternions \#3} & $\begin{bmatrix}\boldsymbol{p}_k\\\boldsymbol{\rho}_k\end{bmatrix}$
                  & $\Exp_{q}(\begin{bmatrix}\boldsymbol{p}_2\\\boldsymbol{\rho}_2\end{bmatrix})\!\ominus\!\Exp_{q}(\begin{bmatrix}\boldsymbol{p}_1\\\boldsymbol{\rho}_1\end{bmatrix})$
                  & $\Log_{q}(\Exp_{q}(\begin{bmatrix}\boldsymbol{p}\\\boldsymbol{\rho}\end{bmatrix})\oplus\mathcal{V}_bh)$ \\
    \midrule
    \textbf{RPY} & $\begin{bmatrix}\boldsymbol{p}_k\\\boldsymbol{\theta}_k\end{bmatrix}$
                  & $\begin{bmatrix}\boldsymbol{p}_2-\boldsymbol{p}_1\\\boldsymbol{\theta}_2-\boldsymbol{\theta}_1\end{bmatrix}$
                  & $\begin{bmatrix}\boldsymbol{p}+\dot{\boldsymbol{p}}h\\\boldsymbol{\theta}+\boldsymbol{W}(\boldsymbol{\theta})\boldsymbol{\omega}_bh\end{bmatrix}$ \\
    % … additional rows …
    \bottomrule
  \end{tabularx}
\end{table}

\textbf{Representations choices:} When using a different floating-base space representation we are faced with the following three main options/decisions: 1) how to select the optimization variables, 2) how to compute differences (e.g. for enforcing integration constraints or costs), and 3) how to perform the integration. We detail most of the possible combinations in Table~\ref{tab:kinematic-ops}\footnote{We denote as $\Exp_{q}(\cdot)$ the operation that transforms a translation vector and a quaternion to the $\SE(3)$ manifold~\cite{sola2018micro}. $\Log_{q}(\cdot)$ is the inverse operation.}: \emph{we focus on the representation of the floating-base}, since we assume that the joint space is a vector space; in other words, we show only the variables related to the floating-base pose representation.

\textbf{Implementation details:} We use semi-implicit Euler integration for all spaces, and rely on the Pinocchio~\cite{carpentier2022} and CasADi~\cite{andersson2019casadi} libraries for acquiring gradients. For the $\SE(3)$ tangent representation we implement all the analytic gradients using the tools provided by Pinocchio. In particular, Pinocchio gives us the Jacobians on the $\SE(3)$ manifold, and we apply the appropriate chain rule to compute the Jacobians for $\mathfrak{se}(3)$. For the Quaternion-based and the Euler Angles representations we utilize the automatic differentiation provided by the CasADi-Pinocchio integration. Finally, as the optimization algorithm, we use the primal-dual interior point method as implemented in the Ipopt solver~\cite{wachter2006_IPOPT}.
The code for the $\SE(3)$-based trajectory optimization and the relevant experiments is available at {\small\url{https://github.com/upatras-lar/se3_trajopt}}.

% \subsection{Transcription based on the $\SE(3)$ Tangent}\label{sec:se3_trajopt}
% \todo{Add stuff}
% %
% \subsection{Transcription based on Quaternions}\label{sec:quat_trajopt}
% \todo{Add stuff}
% %
% \subsection{Transcription based on Euler Angles}\label{sec:rpy_trajopt}
% \todo{Add stuff}
% %
\section{Agile Movement Experiments}\label{sec:agile_exps}
%
% ~\todo{Nice exps here!}
\subsection{Experimental Scenarios}\label{sec:exp_scenarios}
We evaluate five floating-base representations (see Tab.~\ref{tab:kinematic-ops}) across a suite of agile whole-body tasks using three robots: the PAL Robotics Talos humanoid~\cite{stasse2017talos}, the Unitree G1 humanoid, and the Unitree Go2 quadruped.
%All experiments employ inverse-dynamics direct transcription with semi-implicit Euler integration, Ipopt as the NLP solver, and identical time discretization and warm-starting.  
%
Specifically, we test each representation on six motion scenarios, for a total of \textbf{30 optimization runs}.

\noindent\textbf{Tasks:} (see also Fig.~\ref{fig:overview})
\begin{itemize}
  \item \textbf{Talos}:
    \begin{enumerate}
      \item\textbf{Walk}: A Talos humanoid begins from a neutral standing pose and executes a straight‐line walking gait to translate its base $2\,\mathrm{m}$ forward. The contact sequence follows a standard cycle (left‐single support, right‐single support, left‐single support, right‐single support, \dots).
      \item\textbf{Hopscotch}: The Talos humanoid performs a hopscotch pattern over a $2\,\mathrm{m}$ course by alternately hopping twice on one foot. Each step consists of a single‐support contact: first the left foot, then the left foot again, then the right foot, and finally the right foot again before coming to rest in a double support phase.
      \item\textbf{Big jump}: From a stationary stance, Talos jumps forward and lands precisely $1\,\mathrm{m}$ forward. The motion is split into a pre‐launch stance phase, an aerial phase with no contacts, and a two‐foot landing phase.
      \item\textbf{Handstand}: Talos transitions from standing to a handstand and holds the inverted pose in place. The trajectory comprises a crouch‐to‐handstand transfer phase and a phase with both hands in contact before ``coming down'' to a double foot support face again.
    \end{enumerate}
  \item \textbf{Unitree G1}:
    \begin{enumerate}[resume]
      % \item Hopscotch
      \item\textbf{Back-flip}: The Unitree G1 humanoid executes a full backward somersault, landing $0.5\,\mathrm{m}$ behind its takeoff point. The maneuver includes a preparatory phase that ends with a takeoff push, aerial rotation, and impact absorption upon landing.
    \end{enumerate}
  \item \textbf{Unitree Go2}:
    \begin{enumerate}[resume]
      \item\textbf{Side-flip}: A Unitree Go2 quadruped performs a lateral somersault, departing from a neutral stance and landing $0.3\,\mathrm{m}$ sideways. The motion is divided into a preparatory shift, simultaneous hind‐ and fore‐leg launch, aerial flip, and four‐leg landing.
    \end{enumerate}
\end{itemize}
In Table~\ref{tab:costs}, we detail the cost functions and their weights per task, while we also report the discretization timestep and total duration of the behavior. The \emph{Configuration Cost} refers to a regularization cost penalizing the configuration $\boldsymbol{q}_k$ from a reference $\boldsymbol{q}_0$. Similarly, the \emph{Acceleration Cost} refers to a regularization cost penalizing big accelerations: that is, deviations of $\dot{\boldsymbol{v}}_k$ from $\boldsymbol{0}$.

\emph{As evaluation metrics for this scenario,} we report the final cost (if any), number of iterations, number of objective function evaluations, number of Jacobian evaluations (for both the constraints and the objective function), and whether the optimization converged to a solution that solves the task.

\begin{table}[htb]
  \centering
  \caption{Task Parameters.}
  \label{tab:costs}
  \resizebox{\linewidth}{!}{\begin{tabular}{lcccc}
    \toprule
    \textbf{Task} & \textbf{Configuration Cost}~ & ~\textbf{Acceleration Cost} & ~\textbf{Timestep} & ~\textbf{Duration}\\
    \midrule
    Walk (Talos)      & none                        & none             & $0.05\,s$     & $3.3\,s$       \\
    Hopscotch (Talos)    & none                        & none        & $0.05\,s$    & $4.3\,s$             \\
    Big Jump (Talos)      & none                        & none            & $0.05\,s$      & $2.3\,s$       \\
    Handstand (Talos)     & $1\times10^{-9}$           & none         & $0.05\,s$  & $3.4\,s$        \\
    % Hopscotch (G1)       & $1\times10^{-3}$           & $1\times10^{-9}$   & $0.05\,s$    & $3.3\,s$      \\
    Back‐flip (G1)    & $1\times10^{-9}$           & $1\times10^{-6}$     & $0.05\,s$    & $2.4\,s$    \\
    Side‐flip (Go2)       & none                        & none             & $0.02\,s$     & $2.4\,s$       \\
    \bottomrule
  \end{tabular}}
\end{table}

We also assess solver robustness on tasks 1, 3, 5 and 6 by adding zero-mean Gaussian noise of varying standard deviations to the initial warm-start trajectory. \emph{We run 10 replicates} of each variation tuple for \textbf{a total 640 optimization runs}; we have 4 different tasks, 4 different representations (we ignore the ``Quaternion \#3` representation since it was not able to find solutions), and 4 different noise levels. \emph{As an evaluation metric for this scenario,} we report success rates over the 10 replicates and noise levels, and the number of iterations to convergence.

\noindent\textbf{Design criteria.} An ideal trajectory optimizer should require minimal tuning per task or per robot. To this end, across all experiments we:

\begin{itemize}
  \item Use only small regularization costs, with at most minor adjustments between tasks and platforms (see Table~\ref{tab:costs});
  \item Initialize every node in a neutral stance (\textbf{non-informative warm-start});
  \item Avoid robot-specific adaptations to the cost or solver settings;
  \item For the back-flip on G1, we introduce a slightly more informative warm-start: we set the floating-base orientation to the upside-down pose for a few nodes in the middle (in time) of the trajectory. Since we do not use any task-specific cost, this guides the optimizer toward a full somersault rather than a simple backward jump.
\end{itemize}

% To minimize manual tuning while ensuring solver stability, we apply only small configuration‐ and acceleration‐regularization terms. Table~\ref{tab:costs} lists the weights used for each motion scenario. Scenarios marked “none” use zero regularization for both configuration and acceleration costs.

%
\subsection{Experimental Results}\label{sec:results}
\begin{figure}[htbp]
  \centering
  % First row
  \begin{subfigure}[b]{0.77\linewidth}
    \centering
    \includegraphics[width=\linewidth]{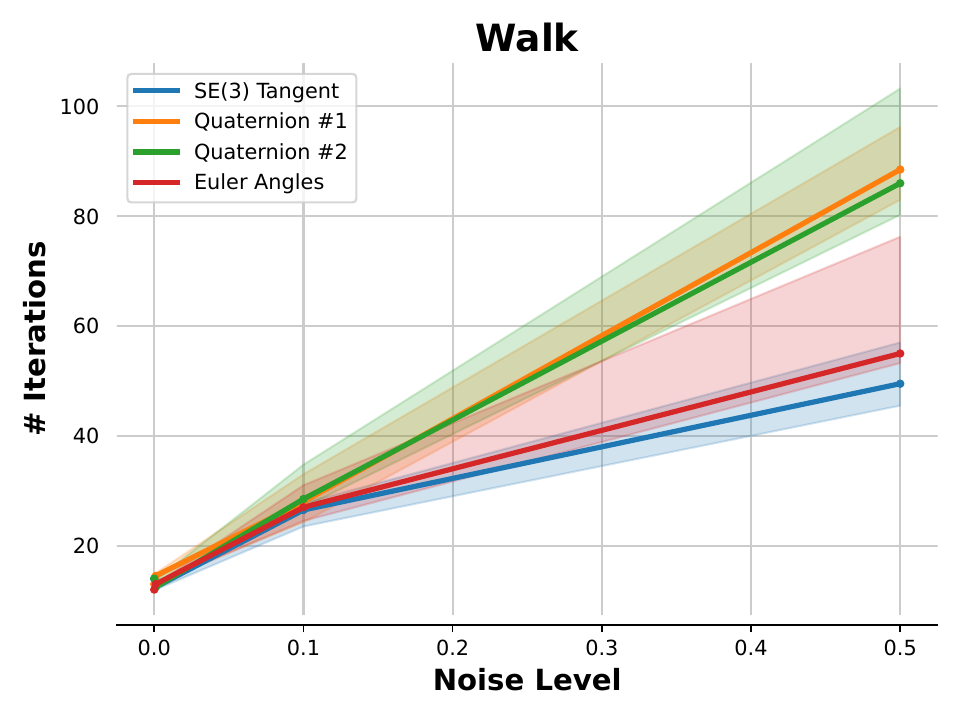}
    \caption{Walk task}
    \label{fig:noise_iter_walk}
  \end{subfigure}
  \hfill
  \begin{subfigure}[b]{0.77\linewidth}
    \centering
    \includegraphics[width=\linewidth]{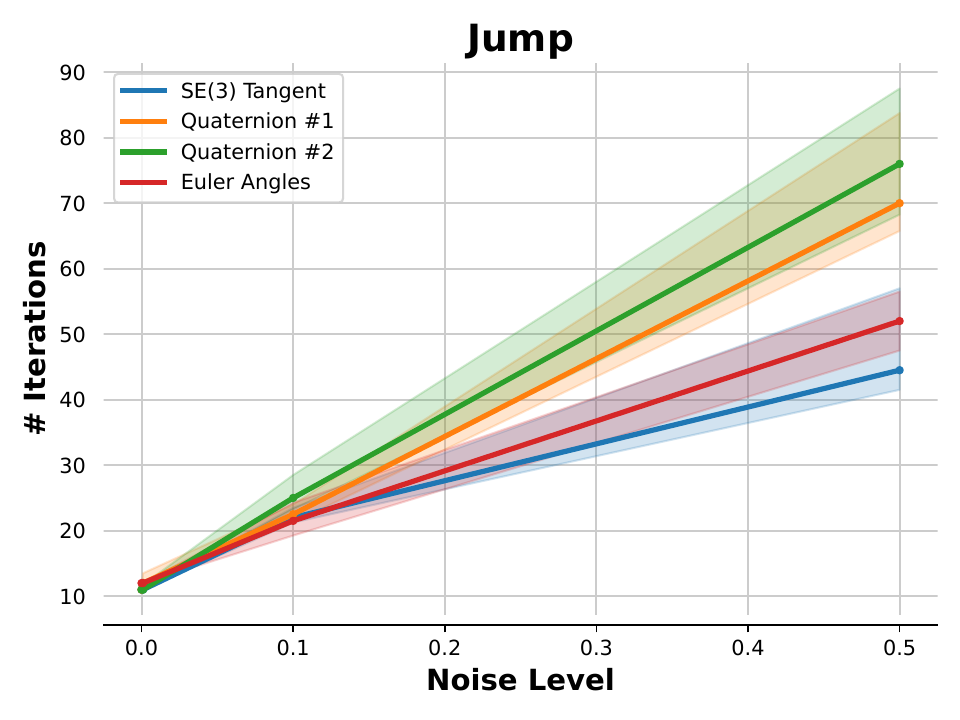}
    \caption{Jump task}
    \label{fig:noise_iter_jump}
  \end{subfigure}

  % \vspace{1em} % small vertical space between rows

  % Second row
  \begin{subfigure}[b]{0.77\linewidth}
    \centering
    \includegraphics[width=\linewidth]{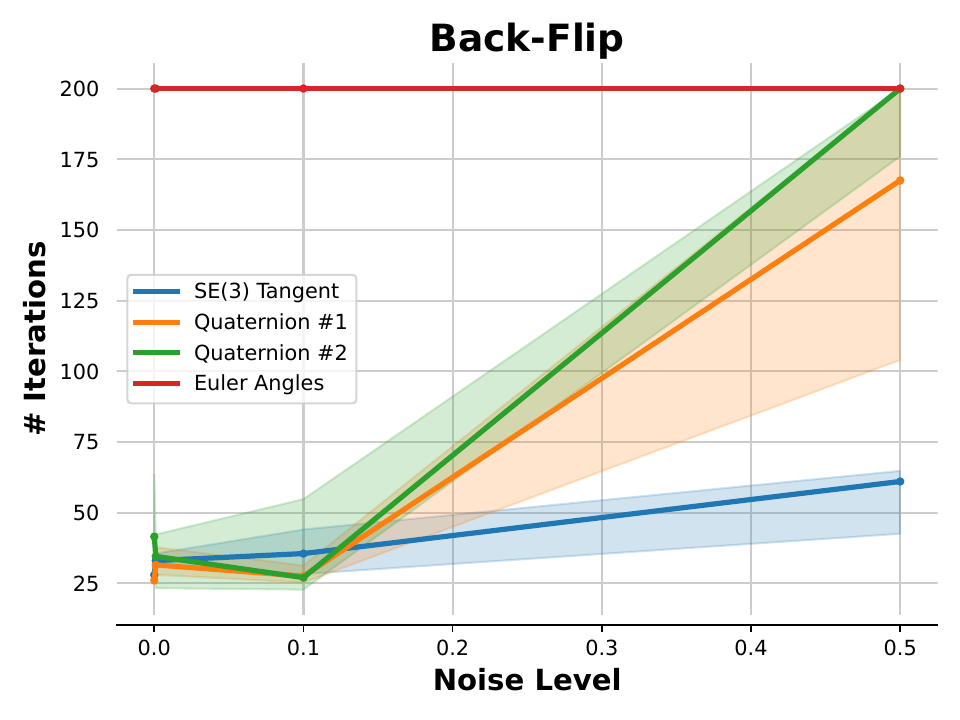}
    \caption{Back-flip task}
    \label{fig:noise_iter_backflip}
  \end{subfigure}
  \hfill
  \begin{subfigure}[b]{0.77\linewidth}
    \centering
    \includegraphics[width=\linewidth]{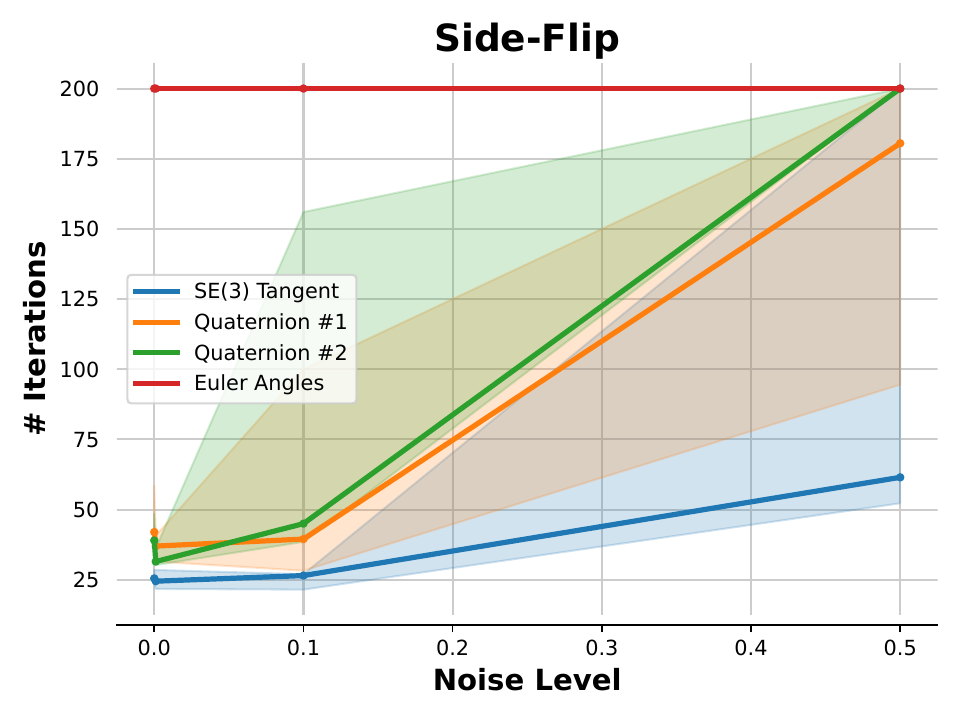}
    \caption{Side-flip task}
    \label{fig:noise_iter_sideflip}
  \end{subfigure}

  \caption{Iterations to convergence versus noise levels. Solid lines are the median over 10 replicates and
the shaded regions are the regions between the 25-th and 75-th percentiles. The $\SE(3)$ tangent representation shows the biggest robustness.}
  \label{fig:noise_iter}
\end{figure}

% The results showcase several interesting outcomes:
% \begin{itemize}
%     \item For tasks where big orientation changes is not required (task 1, 2, 3 4, and 5), most representations (with the exception of ``Quaternion \#3'') converge to similar solutions, and in similar iterations and function/gradient evaluations;
%     \item Combining Quaternions with the manifold operations does not seem a good idea; the representation ``Quaternion \#1'' seem to outperform both Quaternion-manifold mixes (``Quaternion \#2'' and ``Quaternion \#3''), and the representation ``Quaternion \#3'' did not converge in any of our experiments;
%     \item For tasks where changing orientation is important (task 6 and 7), the $\SE(3)$ tangent representation clearly outperforms all other representations;
%     \item The $\SE(3)$ tangent representation is way more robust than other representations when introduced with noise in the warm-start vector, and is able to find agile motions even with big deviations from the original (already non-informative) warm-starting point.
% \end{itemize}
% Overall, the $\SE(3)$ tangent representation seems to perform at least as well or better than all other representations and is able to generate agile motions fast and with simplistic warm-starting.
The results reveal several key insights:
\begin{itemize}
  \item\textbf{Uniform convergence on simple tasks.} For tasks that do not require large orientation changes (Tasks 1–4), all representations ---except for ``Quaternion \#3''--- converge to comparable trajectories in similar numbers of iterations and function evaluations (see Table~\ref{tab:results}).
  \item\textbf{Quaternion–manifold hybrids underperform.} The pure quaternion formulation (``Quaternion \#1'') outperforms both mixed quaternion–manifold variants (``Quaternion \#2'' and ``Quaternion \#3''), and ``Quaternion \#3'' fails to converge in any experiment (see Table~\ref{tab:results}).
  \item\textbf{$\SE(3)$ Tangent-space excels on highly dynamic tasks.} For tasks involving significant reorientation (Tasks 5 and 6), the $\SE(3)$ tangent-space representation clearly outperforms all alternatives in solution quality (see Table~\ref{tab:results}). In particular, in the back- and side- flip tasks, the $\SE(3)$ tangent-space representation is able to converge to highly dynamic behaviors that perform the flips even though the warm-start is mildly informative. On the other hand, quaternion-based representations fail to converge to flipping solutions, and find simple jumping ones.
  \item\textbf{Robustness to initialization noise.} When we add zero-mean Gaussian noise of increasing magnitude to the initial guess, the $\SE(3)$ tangent-space parameterization consistently recovers agile motions (see Table~\ref{tab:results_noise}, Fig.~\ref{fig:noise_iter}). In this experiment, we measure success rate as the ability of the optimizer to converge to a feasible solution, and not the ability to solve the task (we reported this in Table~\ref{tab:results}). For example, in Task 5 and for noise level of $\sigma=10^{-6}$, the ``Quaternion \#1'' variant was converging to a simple jump behavior instead of the flip behaviors, but we report $100\%$ success rate since all runs converged to the jumping behavior.
\end{itemize}

Overall, the $\SE(3)$ tangent-space representation matches or exceeds the performance of all other floating-base parameterizations. Nevertheless, ``Quaternion \#1` is a competitive alternative if one does not need very agile and dynamic movements or large re-orientations.

% \textbf{Remarks.}
\textbf{Qualitative analysis.} Fig.~\ref{fig:overview} shows a few behaviors generated by the $\SE(3)$ Tangent space representation. In the \emph{supplementary video} (also available at \url{https://lar.upatras.gr/projects/ibrics.html}), you can find more informative visualizations of the generated behaviors.

\begin{table}[tbh]
  \centering
  % \vspace{-1em}
  \caption{Success rates under varying warm‐start noise levels.}
  \label{tab:results_noise}
  \resizebox{\linewidth}{!}{\begin{tabular}{llcccc}
    \toprule
    \textbf{Task} & \textbf{Representation} & \(\sigma=10^{-6}\) & \(\sigma=10^{-3}\) & \(\sigma=0.1\) & \(\sigma=0.5\)\\
    \midrule
    \multirow{5}{*}{Walk (Talos)} 
      & Euler angles         & $100\%$ & $100\%$ & $100\%$ & $100\%$\\
      & Quaternion \#1       & $100\%$ & $100\%$ & $100\%$ & $100\%$\\
      & Quaternion \#2       & $100\%$ & $100\%$ & $100\%$ & $100\%$\\
      % & Quaternion \#3       & $0\%$ & $0\%$ & $0\%$ & \texttt{…/10}\\
      & $\SE(3)$ Tangent  & $100\%$ & $100\%$ & $100\%$ & $100\%$\\
    \addlinespace
    \midrule
    \multirow{5}{*}{Big jump (Talos)} 
      & Euler angles         & $100\%$ & $100\%$ & $100\%$ & $100\%$\\
      & Quaternion \#1       & $100\%$ & $100\%$ & $100\%$ & $100\%$\\
      & Quaternion \#2       & $100\%$ & $100\%$ & $100\%$ & $90\%$\\
      % & Quaternion \#3       & $0\%$ & $0\%$ & $0\%$ & \texttt{…/10}\\
      & $\SE(3)$ Tangent  & $100\%$ & $100\%$ & $100\%$ & $100\%$\\
    \addlinespace
    \midrule
    \multirow{5}{*}{Back‐flip (G1)} 
      & Euler angles         & $0\%$ & $20\%^*$ & $0\%$ & $0\%$\\
      & Quaternion \#1       & $100\%^*$ & $100\%^*$ & $100\%^*$ & $60\%^*$\\
      & Quaternion \#2       & $100\%^*$ & $100\%^*$ & $100\%^*$ & $40\%^*$\\
      % & Quaternion \#3       & \texttt{…/10} & \texttt{…/10} & \texttt{…/10} & \texttt{…/10}\\
      & $\SE(3)$ Tangent  & $100\%$ & $100\%$ & $100\%$ & $100\%$\\
    \addlinespace
    \midrule
    \multirow{5}{*}{Side‐flip (Go2)} 
      & Euler angles         & $0\%$ & $0\%$ & $0\%$ & $0\%$\\
      & Quaternion \#1       & $100\%^*$ & $100\%^*$ & $80\%^*$ & $50\%^*$\\
      & Quaternion \#2       & $100\%^*$ & $100\%^*$ & $80\%^*$ & $40\%^*$\\
      % & Quaternion \#3       & \texttt{…/10} & \texttt{…/10} & \texttt{…/10} & \texttt{…/10}\\
      & $\SE(3)$ Tangent  & $100\%$ & $100\%$ & $100\%$ & $60\%$\\
    \bottomrule\\
    \multicolumn{5}{l}{\footnotesize\emph{$^*$ Converged but did not find the flip behavior.}}
    \vspace{-2em}
  \end{tabular}}
\end{table}

\begin{table*}[htb]
  % \small
  \normalsize
  \centering
  \caption{Optimization results for five floating‐base representations across six tasks.}
  \label{tab:results}
  \begin{tabular}{llccccc}
    \toprule
    \textbf{Task} & ~\textbf{Representation}~ & ~\textbf{Cost}~ & ~\textbf{Iterations}~ & ~\textbf{Obj. evals}~ & ~\textbf{Jac. evals}~ & ~\textbf{Success}~ \\
    \midrule
    \multirow{5}{*}{Walking (Talos)} 
      & Euler angles         & \texttt{-} & \texttt{12} & \texttt{13} & \texttt{42} & \texttt{Yes} \\
      & Quaternion \#1       & \texttt{-} & \texttt{13} & \texttt{15} & \texttt{42} & \texttt{Yes} \\
      & Quaternion \#2       & \texttt{-} & \texttt{14} & \texttt{16} & \texttt{45} & \texttt{Yes} \\
      & Quaternion \#3       & \texttt{-} & \texttt{-} & \texttt{-} & \texttt{-} & \texttt{No} \\
      & $\SE(3)$ Tangent  & \texttt{-} & \texttt{14} & \texttt{16} & \texttt{45} & \texttt{Yes} \\
    \addlinespace
    \midrule
    \multirow{5}{*}{Hopscotch (Talos)} 
      & Euler angles         & \texttt{-} & \texttt{14} & \texttt{18} & \texttt{45} & \texttt{Yes} \\
      & Quaternion \#1       & \texttt{-} & \texttt{12} & \texttt{13} & \texttt{39} & \texttt{Yes} \\
      & Quaternion \#2       & \texttt{-} & \texttt{16} & \texttt{17} & \texttt{51} & \texttt{Yes} \\
      & Quaternion \#3       & \texttt{-} & \texttt{-} & \texttt{-} & \texttt{-} & \texttt{No} \\
      & $\SE(3)$ Tangent  & \texttt{-} & \texttt{10} & \texttt{14} & \texttt{33} & \texttt{Yes} \\
    \addlinespace
    \midrule
    \multirow{5}{*}{Big jump (Talos)} 
      & Euler angles         & \texttt{-} & \texttt{12} & \texttt{13} & \texttt{39} & \texttt{Yes} \\
      & Quaternion \#1       & \texttt{-} & \texttt{11} & \texttt{12} & \texttt{36} & \texttt{Yes} \\
      & Quaternion \#2       & \texttt{-} & \texttt{11} & \texttt{12} & \texttt{36} & \texttt{Yes} \\
      & Quaternion \#3       & \texttt{-} & \texttt{-} & \texttt{-} & \texttt{-} & \texttt{No} \\
      & $\SE(3)$ Tangent  & \texttt{-} & \texttt{11} & \texttt{12} & \texttt{36} & \texttt{Yes} \\
    \addlinespace
    \midrule
    \multirow{5}{*}{Handstand (Talos)} 
      & Euler angles         & \texttt{3.710e-7} & \texttt{53} & \texttt{68} & \texttt{162} & \texttt{Yes} \\
      & Quaternion \#1       & \texttt{3.71e-7} & \texttt{58} & \texttt{58} & \texttt{135} & \texttt{Yes} \\
      & Quaternion \#2       & \texttt{3.81e-7} & \texttt{96} & \texttt{96} & \texttt{159} & \texttt{Yes} \\
      & Quaternion \#3       & \texttt{-} & \texttt{-} & \texttt{-} & \texttt{-} & \texttt{No} \\
      & $\SE(3)$ Tangent  & \texttt{3.91e-7} & \texttt{88} & \texttt{104} & \texttt{267} & \texttt{Yes} \\
    \addlinespace
    \midrule
    % \multirow{5}{*}{Hopscotch (G1)} 
    %   & Euler angles         & \texttt{…} & \texttt{…} & \texttt{…} & \texttt{…} & \texttt{Yes/No} \\
    %   & Quaternion \#1       & \texttt{…} & \texttt{…} & \texttt{…} & \texttt{…} & \texttt{Yes/No} \\
    %   & Quaternion \#2       & \texttt{…} & \texttt{…} & \texttt{…} & \texttt{…} & \texttt{Yes/No} \\
    %   & Quaternion \#3       & \texttt{…} & \texttt{…} & \texttt{…} & \texttt{…} & \texttt{Yes/No} \\
    %   & $\SE(3)$ Tangent  & \texttt{…} & \texttt{…} & \texttt{…} & \texttt{…} & \texttt{Yes/No} \\
    % \addlinespace
    % \midrule
    \multirow{5}{*}{Back-flip (G1)} 
      & Euler angles         & \texttt{12.5} & \texttt{200} & \texttt{414} & \texttt{603} & \texttt{No} \\
      & Quaternion \#1       & \texttt{0.1} & \texttt{26} & \texttt{30} & \texttt{81} & \texttt{No$^*$} \\
      & Quaternion \#2       & \texttt{0.1} & \texttt{71} & \texttt{84} & \texttt{216} & \texttt{No$^*$} \\
      & Quaternion \#3       & \texttt{-} & \texttt{-} & \texttt{-} & \texttt{-} & \texttt{No} \\
      & $\SE(3)$ Tangent  & \texttt{5.2e-5} & \texttt{28} & \texttt{47} & \texttt{87} & \texttt{Yes} \\
    \addlinespace
    \midrule
    \multirow{5}{*}{Side-flip (Go2)} 
      & Euler angles         & \texttt{-} & \texttt{200} & \texttt{632} & \texttt{603} & \texttt{No} \\
      & Quaternion \#1       & \texttt{-} & \texttt{39} & \texttt{45} & \texttt{120} & \texttt{No$^*$} \\
      & Quaternion \#2       & \texttt{-} & \texttt{56} & \texttt{316} & \texttt{171} & \texttt{No$^*$} \\
      & Quaternion \#3       & \texttt{-} & \texttt{-} & \texttt{-} & \texttt{-} & \texttt{No} \\
      & $\SE(3)$ Tangent  & \texttt{-} & \texttt{29} & \texttt{50} & \texttt{90} & \texttt{Yes} \\
    \bottomrule\\
    \multicolumn{7}{l}{\footnotesize\emph{$^*$ The solver converged to a feasible solution, but it was not performing the flip. See also the supplementary video.}}
  \end{tabular}
  \vspace{-2em}
\end{table*}

\section{Conclusion}\label{sec:conclusion}
In this work, we presented a systematic comparison of five floating‐base parameterizations, based on Euler angles, quaternions, and the $\SE(3)$ tangent‐space, in the context of agile whole‐body motion planning for legged robots. Using identical transcription settings, solver parameters, and non‐informative warm starts across six benchmark tasks, our experiments demonstrated that:
\begin{itemize}
  \item All vector representations (Euler, Quaternion \#1) perform similarly on low‐rotation tasks (Task 1-4), but require more careful handling in high-rotation tasks (Task 5-6): Euler can fail to converge, while Quaternions might converge to suboptimal solutions.
  \item Quaternion–manifold hybrids underperform or even fail to converge.
  \item The $\SE(3)$ tangent‐space formulation consistently yields the highest solution quality on dynamic movements that require large re-orientations.
  \item $\SE(3)$ tangent‐space is robust to perturbations in the initial guess, recovering agile behaviors under noisy initializations.
\end{itemize}
Overall, the $\SE(3)$ tangent‐space representation provides the best balance between expressive power and compatibility with mature NLP solvers (e.g., Ipopt), enabling fast, reliable trajectory optimization for challenging legged‐robot tasks.

Building on the promising performance of the $\SE(3)$ tangent‐space formulation, our next step is to validate whether the produced behaviors can run on physical hardware. We are confident that this is possible given that simpler models have been deployed in real systems~\cite{winkler2018gait,ding2021representation}. In parallel, we plan to implement $\SE(3)$-based Sequential Quadratic Programming (SQP) solvers dedicated to optimal control problems, which will allow us to run fast real-time Model Predictive Control (MPC) schemes (e.g. based on recent ADMM-based methods~\cite{jordana2023stagewise}); we are also planning to incorporate learned dynamics models~\cite{khadivar2023self,chatzilygeroudis2017black} to facilitate real-world deployment.

% We also intend to explore hybrid optimization strategies that embed manifold retraction operators directly within off‐the‐shelf NLP solvers, potentially enhancing convergence on manifold variables without sacrificing solver maturity.
Finally, we aim to incorporate learning‐based methods ---e.g., leveraging demonstration data~\cite{totsila2025sensorimotor} or reinforcement‐learning policies for better warm-starting~\cite{melon2020reliable,chatzilygeroudis2019survey} or the other way around~\cite{bogdanovic2022model}--- to reduce solve times, and generalize our approach beyond legged robots to other systems, such as multi‐link manipulators and aerial manipulation platforms.

% \renewcommand*{\thefootnote}{\fnsymbol{footnote}}
% \setcounter{tempcounter}{0}
% \addtocounter{tempcounter}{\value{footnote}}
% \setcounter{footnote}{1}
% \footnotetext{It did not run}
% \renewcommand*{\thefootnote}{\arabic{footnote}}
% \setcounter{tempcounter}{\value{tempcounter}}

% \addtolength{\textheight}{-11cm}   % This command serves to balance the column lengths

%%%%%%%%%%%%%%%%%%%%%%%%%%%%%%%%%%%%%%%%%%%%%%%%%%%%%%%%%%%%%%%%%%%%%%%%%%%%%%%%

%%%%%%%%%%%%%%%%%%%%%%%%%%%%%%%%%%%%%%%%%%%%%%%%%%%%%%%%%%%%%%%%%%%%%%%%%%%%%%%%

%%%%%%%%%%%%%%%%%%%%%%%%%%%%%%%%%%%%%%%%%%%%%%%%%%%%%%%%%%%%%%%%%%%%%%%%%%%%%%%%
% \section*{Appendix}


% \section*{Acknowledgment}

% If any.

%%%%%%%%%%%%%%%%%%%%%%%%%%%%%%%%%%%%%%%%%%%%%%%%%%%%%%%%%%%%%%%%%%%%%%%%%%%%%%%%
% \vspace{-0.5em}
\bibliographystyle{ieeetr}
\bibliography{references}

\end{document}